\documentclass[]{interact}

\usepackage{epstopdf}
\usepackage[caption=false]{subfig}

\usepackage[numbers,sort&compress]{natbib}
\bibpunct[, ]{[}{]}{,}{n}{,}{,}
\renewcommand\bibfont{\fontsize{10}{12}\selectfont}
\makeatletter
\def\NAT@def@citea{\def\@citea{\NAT@separator}}
\makeatother

\theoremstyle{plain}

\theoremstyle{definition}

\theoremstyle{remark}

\usepackage{amsmath,amssymb,amsfonts}
\usepackage{algorithmic}
\usepackage{graphicx}
\graphicspath{{illustration/}}
\usepackage{textcomp}
\usepackage{xcolor}
\usepackage{makecell}
\usepackage{array} 
\usepackage{stfloats}
\usepackage{geometry}
\usepackage{comment}

\begin{document}


\title{
	DeFNet: Deconstructed Strategy for Multi-step Fabric Folding Tasks
}

\author{
\name{Ningquan Gu$^1$, Ruhan He$^1$\textsuperscript{*}\thanks{\textsuperscript{*}CONTACT Ruhan He. Email: heruhan@wtu.edu.cn} and Lianqing Yu$^2$}
\affil{ $^1$School of Computer Science and Artificial Intelligence, Wuhan Textile University, China
}
\affil{ $^2$School of Mechanical Engineering and Automation, Wuhan Textile University, China
}
}

\maketitle

\begin{abstract}
Fabric folding through robots is complex and challenging due to the deformability of the fabric. Based on the deconstruction strategy, we split the tough multi-step fabric folding task into three relatively simple sub-tasks and propose a Deconstructed Fabric Folding Network (DeFNet), including three corresponding modules to solve them. (1) We use the Folding Planning Module (FPM),  which is based on the Latent Space Roadmap, to infer the shortest folding intermediate states from the start to the goal in latent space. (2) We utilize the flow-based approach, Folding Action Module (FAM), to calculate the action coordinates and execute them to reach the inferred intermediate state. (3) We introduce an Iterative Interactive Module (IIM) for multi-step fabric folding tasks, which can iteratively execute the FPM and FAM after every grasp-and-place action until the fabric reaches the goal. Experimentally, we demonstrate our method on fabric folding tasks against three baselines in simulation. We also apply the method to an existing robotic system and present its performance. 
\end{abstract}


\section{Introduction}
Fabric manipulation has many applications in daily life and industry. However, it is complex and challenging for robots to realize functions like picking the target fabric from a stack of clothes, unfolding towels, and laundry folding or smoothing  \cite{maitin2010cloth,miller2012geometric,shibata2012trajectory,seita2019deep}. 
In the grasping and placing process, self-occlusions in deformable configurations  \cite{huang2022mesh} lead to only partial observability  \cite{kaelbling2013integrated}, which increases the complexity compared to the rigid objects. 

Early methods for folding or smoothing fabric mainly relied on non-learned methods with pre-set actions \cite{maitin2010cloth,bersch2011bimanual}. However, these policies were often slow and lacked the ability to generalize to arbitrary initial and target states. 
In recent years, researchers have explored learning-based approaches for fabric manipulation \cite{tsurumine2019deep,ha2022flingbot,seita2019deep,canberk2022cloth}, which have shown improved results compared to traditional non-learned methods.
In particular, the use of low-dimensional latent space representations for planning has gained popularity in fabric manipulation tasks due to their ability to address the complex dynamics of the state space, enabling the consideration of states that would otherwise be difficult to handle.
For instance, Lippi et al.  \cite{lippi2020latent,lippi2022enabling} proposed the Visual Action Planning (VAP) approach, which utilized the Latent Space Roadmap (LSR) to generate a series of images that depict a plan visually in sequential order and employed the Action Proposal Module (APM) to predict the actions between them in box stacking and T-shirt manipulation tasks.
However, their action proposal policy directly outputted both the grasp and the place coordinate points simultaneously. It is difficult to capture the underlying inherent complexity and interdependence of the action space, as the place point is heavily dependent on the grasp point \cite{wu2019learning}.
In addition, Weng et al.  \cite{weng2022fabricflownet} applied FlowNet to fabric manipulation tasks, allowing them to reason about the flow of fabric and obtain more precise grasp-and-place points. Their approach demonstrated significant outperformance compared to cutting-edge model-free and model-based fabric manipulation policies. However, a limitation of their policy is that it requires knowledge of each sub-goal in the multi-step task, which can be challenging in specific scenarios.

\begin{figure}[tbp]
	\centerline
	{
		\includegraphics[
		width=1.0\textwidth]{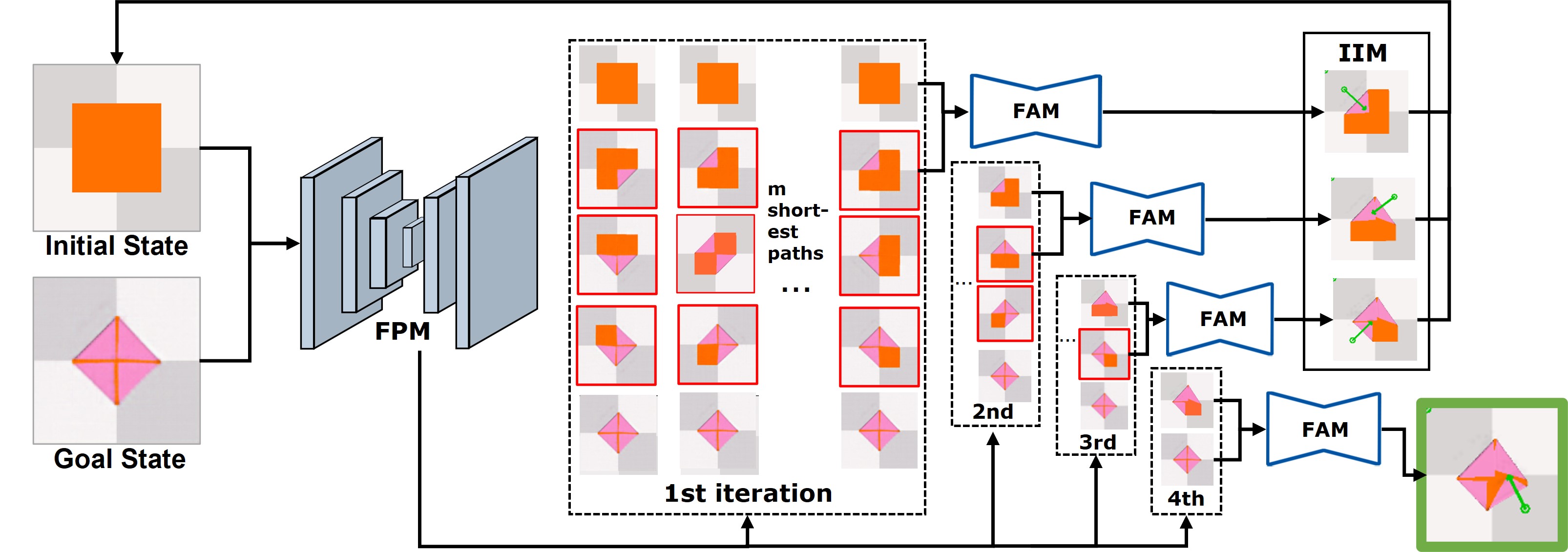}
	}
	\caption{
		Our proposed approach for fabric folding deconstructs the task into three modules. Firstly, the Folding Planning Module (FPM) reasons about the intermediate folding states of the fabric, given the initial and the goal states. The FPM generates a sequence of images with red borders, representing the inferred intermediate states. For multi-folding tasks, there are typically several shortest folding paths. Secondly, the Folding Action Module (FAM) utilizes a flow-based approach. We randomly select one of the shortest paths and feed the corresponding images into the FAM to obtain the grasp-and-place actions between the intermediate folding states. Finally, the Iterative Interactive Module (IIM) carries out each grasp-and-place action (indicated by a green arrow on the images) and takes the current observation of the fabric as a new start state to input to the FPM again. The IIM iteratively executes the first and second modules after each grasp-and-place action until the goal state is reached (the image with a green border). By utilizing these three modules, our approach can efficiently and effectively solve complex fabric folding tasks.		
	}
	\label{fig-framework}
\end{figure}


To address the complexity of fabric folding tasks, we propose the Deconstructed Fabric Folding Network (DeFNet) that utilizes a deconstruction strategy to split the task into three relatively simple sub-tasks. The framework of DeFNet is shown in Figure~\ref{fig-framework}.

DeFNet consists of three modules. The first module is the Folding Planning Module (FPM), which operates in latent space to determine the fabric folding paths. By taking the initial and goal states as input, FPM generates a straightforward sequence of intermediate states. We utilize the Latent Space Roadmap (LSR)  \cite{lippi2022enabling} as the planning policy within FPM.
The second module is the Folding Action Module (FAM), which employs an optical flow policy to calculate the grasp-and-place points between each pair of intermediate fabric states. This module enables precise determination of the actions required for the folding process.
The third module is the Iterative Interactive Module (IIM), which plays a crucial role in enhancing the execution process. After each grasp-and-place action, IIM utilizes the current fabric observation as a new start state and feeds it back into the FPM. By iteratively executing the first and second modules after each action, IIM effectively reduces potential uncertainties in the execution process and facilitates more reliable perception and planning.
In our experiments, our approach outperforms state-of-the-art fabric manipulation policies.

The paper's contributions can be summarized as follows:
\begin{itemize}
	\item Based on the deconstruction strategy, we propose a novel framework DeFNet to split the complex folding task into three relatively simple sub-tasks, which makes it possible to improve the accuracy of each folding step in the long-horizon sequential folding task and increase the overall accuracy.
	\item We design an Iterative Interactive Module for multi-step fabric folding tasks, which integrates FPM and FAM into our framework perfectly. Meanwhile, the iterative module is very beneficial in enhancing the accuracy of each step.
	\item We experimentally evaluate DeFNet against three SOTA methods in simulation and perform the ablations, which shows the best performance.
\end{itemize}

\section{Related work}
\subsection{Learning for fabric manipulation}
Fabric manipulation by robots is a challenging task due to the high dimensional space of fabric  \cite{tsurumine2019deep}, and representing the state of the fabric is difficult. 
There are several unsolved and under-optimized tasks in the field of fabric manipulation, including fabric simulation \cite{lin2021softgym}, fabric smoothing \cite{seita2020deep}, bed making \cite{seita2019deep}, and fabric folding \cite{ganapathi2021learning}. 
Many prior works have used traditional computer vision algorithms to describe the fabric and pre-set trajectories to manipulate it. However, these approaches have limitations in handling complex configurations of fabric and may not be effective in solving long-horizon sequential folding tasks.
For instance, traditional computer vision algorithms, including using Harris corner detection and discontinuity to get the peak ridges and peak corners  \cite{willimon2011model}, and using RANSAC to locate the fabric corners \cite{maitin2010cloth}, or even lifting the cloth and flinging it to obtain a new goal shape \cite{triantafyllou2011vision}, tend to fail under complex configurations of fabric.
Applying learning methods, especially deep learning networks or deep reinforcement learning, provides more options for fabric manipulation tasks. For instance, Lee et al.  \cite{lee2021learning} learned a model-free algorithm for folding fabric by collecting a dataset through the interaction between a real robot and the fabric. However, their performance is limited by the discrete action space, and they are not very effective at solving long-horizon sequential folding tasks. 
Similarly, Hoque et al.  \cite{hoque2020visuospatial} learned a visual dynamics model and used CEM to plan, which can be used to solve arbitrary goals and long-horizon tasks, but their approach requires a significant number of grasp-and-place actions in fabric manipulation tasks, potentially requiring multiple actions to fold one corner.
In contrast, our framework can deal with the long-horizon sequential folding task, which infers the intermediate folding states with the shortest folding paths.

\subsection{Latent space roadmap}
Approaches to planning in complex scenes, where the system state cannot be analytically determined, can be identified into two main strategies according to where the planning is executed.
The first is high-dimensional image space. Finn et al. in 2017  \cite{finn2017deep} predicted stochastic pixel flow from frame to frame using Long-Short Term Memory blocks. 
Hoque et al.  \cite{ganapathi2021learning} introduced VisuoSpatial Foresight in 2021 as a method to predict fabric movement. This approach is based on previous research, which involved learning the visual dynamics of domain-randomized RGB images and depth information simultaneously.
In particular, Nair et al.  \cite{nair2017combining} utilized five hundred hours to collect data to learn a rope's inverse dynamics.
Alternatively, planning in low-dimensional latent state space can decrease the complexity of the input image space.
In 2020, Lippi et al. \cite{lippi2020latent} proposed and implemented a Visual Action Planning(VAP), a graph-based structure that globally captures the latent system dynamics. It can predict the state of rigid objects and deformable objects through the Latent Space Roadmap(LSR) module of VAP. In 2022, Lippi et al. \cite{lippi2022enabling} improved the performance of the LSR module, including improving the extension of the LSR building algorithm and adding the dynamic adjustment of the critical hyperparameter.
But, their action proposal policy directly outputted both the grasp and the place locations simultaneously. 
It is difficult to capture the underlying inherent complexity and interdependence of the action space, as the place point is heavily dependent on the grasp point  \cite{wu2019learning}.
Our method differs from previous representations in that it uses a flow-based approach, which has been shown to outperform previous methods for goal-based fabric manipulation significantly.

\subsection{Optical flow for policy learning}
Estimating per-pixel correspondences between two images can get the optical flow of each pixel. 
In 2015, Dosovitskiy et al. \cite{dosovitskiy2015flownet} proposed a FlowNet into the video, which can predict optical flow at ten image pairs per second, achieving excellent accuracy among real-time methods. 
In 2022, Weng et al. \cite{weng2022fabricflownet} applied the FlowNet to fabric manipulation, reasoning the flow of fabric in manipulation and getting more precise grasp-and-place points, demonstrating a significant outperforming cutting-edge fabric manipulation policy.
However, the policy FabricFlowNet (FFN) of Weng et al., must know each intermediate folding sub-state of the fabric. 
In many cases, we often only have the initial and goal states in the multi-step folding tasks.
Instead, our method needs fewer inputs, just the initial and goal states, without inputting the sub-goals. 

\section{Problem statement} \label{problem_statement}
We aim to enable a robot to perform multi-step folding tasks.
The tasks can be formulated as follows: given a start and goal states of fabric, after a series of dynamically valid actions, we can get the final state.

The problem is formalized as follows:
The input is a tuple $ \langle I_{\text {start}}, I_{\text {goal}} \rangle $.
The $I_{\text {start}}$ is a flattened fabric. $I_{\text {goal}}$ represents the folded goal state which we want to reach. 

The $a_i\in\mathcal{U}$ represents the action that the robot executed.
$\mathcal{U}$ represents actions space.
When we execute the action $a_i$, $a_i\in\mathcal{A}$, the fabric can change the current state ${I}_{\text {i}}$  to the next state ${I}_{\text {i+1}}$. 
After a series of action, $\mathcal{A}: \langle a_0,a_1,\cdots,a_n \rangle $, we can get the final state $I_{final}$ of the fabric. 
\begin{equation}
	(\mathcal{A},I_{final})=\pi(I_{\text {start}},I_{\text {goal}})
\end{equation}
We are interested in learning the $\pi$ to minimize the number of actions and the difference between the $I_{\text {goal}}$ and $I_{\text {final}}$.

\section{Approach}
\subsection{Method overview}
Our novel strategy is to deconstruct the folding task. As Figure~\ref{fig-framework} presents, we can improve the performance by splitting the task into three relatively simple sub-tasks: firstly, we analyze the connections between the start and goal states to obtain the shortest intermediate folding states of the fabric. Second, we compute the grasp-and-place action points between the initial and predicted intermediate folding states. Third, after executing each folding action, we take the current observation of the fabric as a new start state to input the first module iteratively.

We factor the policy as:
\begin{equation}
	\pi =\pi_{\text {p}} \cdot \pi_{\text {a}}
\end{equation}

\begin{figure}[htbp]
	\centerline
	{
		\includegraphics[
		width=0.8\textwidth]{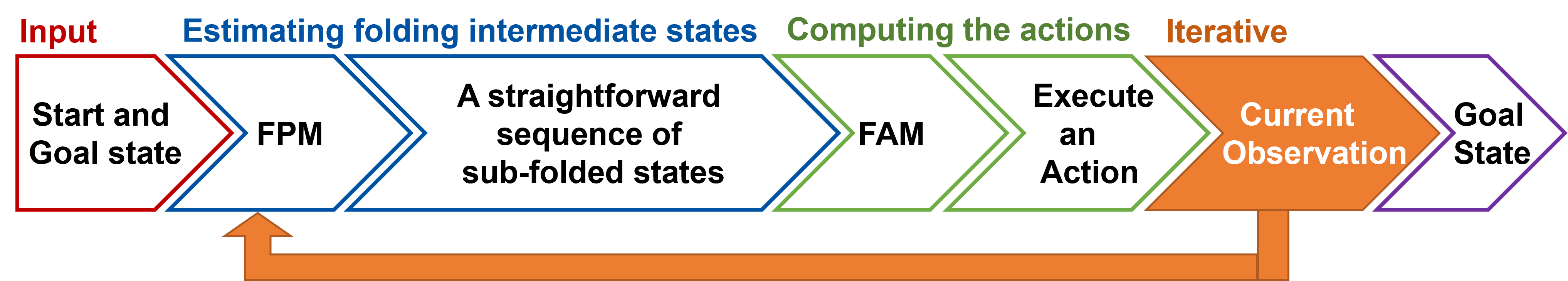}
	}
	\caption{The pipeline of our approach}
	\label{fig-pipeline}
\end{figure}

Figure~\ref{fig-pipeline} provides the overall pipeline of our approach. In the fabric folding task, our framework takes in a pair of tuples (the red color arrow box), $\langle I_{\text {start}}, I_{\text {goal}} \rangle$.
The FPM (the blue color arrow box) utilizes the low-dimensional spaces to predict the intermediate folding states, generating a straightforward sequence of sub-steps $\mathcal{G}$: $\langle I_{\text {0}} =I_{\text {start}}, \hat{I}_{\text {1}},\hat{I}_{\text {2}}\cdots\hat{I}_{\text {n-1}}, I_{\text {n}} =I_{\text {goal}} \rangle$. 
The path from the start to the goal is the shortest, meaning it only needs n times of folding action if we want the goal shape. There may be several shortest folding paths, and we can randomly select a path without affecting the result. 
Next, we use the FAM to compute the grasp-and-place points based on the outputs of the FPM. 
After each folding action, the fabric enters a new state, and we iteratively input the current observation into the FPM until the goal state is achieved.
\subsection{Folding planning module}
The Folding Planning Module (FPM) learns a policy $\pi_\text{p}$ to infer the shortest intermediate folding states $\mathcal{G}$ from $I_{\text {start}}$ to $I_{\text {goal}}$, with the target of minimizing the number of folding steps required.
\begin{equation}
	\mathcal{G}=\pi_\text{p}\left(I_{\text {start}}, I_{\text {goal}}\right)
\end{equation}


\begin{figure}[tbp]
	\centerline
	{
		\includegraphics[
		width=1.0\textwidth]{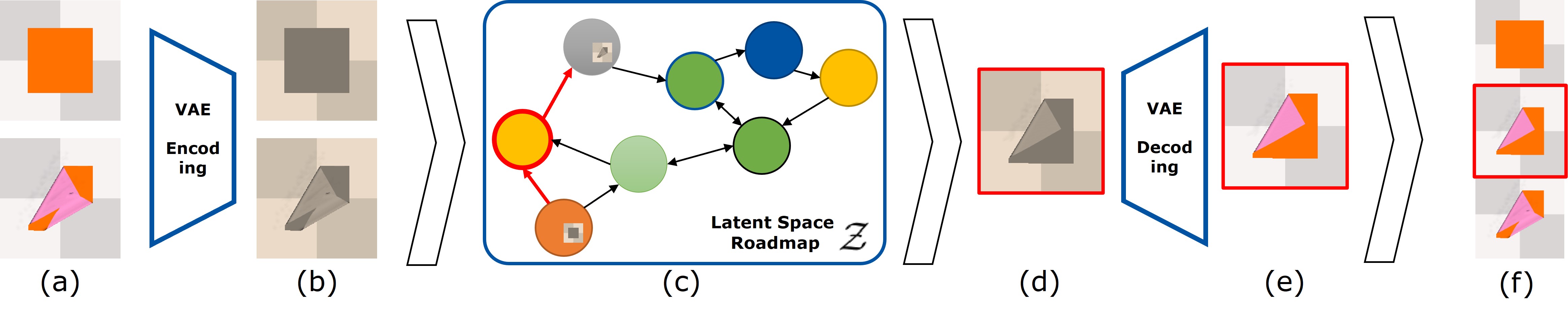}
	}
	\caption{
		The Folding Planning process consists of several steps. First, the Variational Autoencoder (VAE) transforms the input images (a) into a low-dimensional latent representation (b). 
		In this low-dimensional space, the manipulation planning is executed in a built Latent Space RoadMap (c), which determines the optimal sequence of intermediate folding states to reach the goal state. 
		Finally, the predicted results (d) are transformed back into the high-dimensional image space (e) using the VAE. 
		By combining the high-dimensional predicted results with the initial and goal states, the complete shortest folding plan steps are obtained. 
	}
	\label{fig-visualplan}
\end{figure}

We implement the Folding Planning Module by referencing a part of Lippi et al.'s work, which consists of two components: the Variational Autoencoder (VAE) \cite{kingma2013auto}, and the Latent State Roadmap (LSR)  \cite{lippi2022enabling}. 
The VAE maps observations, which are provided in the form of high-dimensional states, to extract the corresponding structured low-dimensional states in latent space, and it can also generate observations based on the latent state. 
The LSR is built in the latent space $\mathcal{Z}$ obtained from the VAE. 
The LSR constructs and links clusters of similar states to identify the latent plan paths between the start and goal states, which have been extracted by the VAE. This enables the identification of latent plans between the extracted states $z_{\text {start}}$ and $z_{\text {goal}}$. Finally, the LSR is used to retrieve all shortest paths between the two identified states, allowing us to identify the shortest intermediate folding states required to reach the goal state.

When performing the Folding Planning Module, we start with the initial and goal states, a trained VAE, and a built LSR. 
As shown in Figure~\ref{fig-visualplan}, the initial state (a) is first encoded by the VAE into a low-dimensional latent representation, shown as the brown images in (b). 
The encoded information is then mapped into the LSR to find the corresponding nodes. 
Next, we identify all of the shortest paths in the LSR between these nodes, shown as the red line in (c). 
From these paths, we select the intermediate node, indicated as an orange node with a red circle. 
The latent information of the intermediate node is shown in (d). 
Finally, the latent information in (d) is decoded by the VAE to generate the visual image in (e). 
By combining the planned intermediate folding states with the start and goal states, we obtain the complete shortest folding plan steps.
To determine how to reach the intermediate folding sub-steps through grasping and placing, we use a Flow-based policy to calculate the grasp-and-place points, as described in Section~\ref{fam}. 
This policy allows the robot to execute the planned folding sub-steps by identifying the optimal grasping and placing points for the fabric during each step of the folding process.


The loss function of the Folding Planning Module (FPM) is influenced by two factors: the Variational Autoencoder (VAE) loss $\mathcal{L}_{\text {vae}}$ and the action term $\mathcal{L}_{\text {action}}$, as described in  \cite{lippi2022enabling}.
Then, the loss of FPM can be defined as that:
\begin{equation}
	\mathcal{L}=\frac{1}{2}\left(\mathcal{L}_{\text {vae}}\left(I_{1}\right)+\mathcal{L}_{\text {vae}}\left(I_{2}\right)\right)+\alpha \cdot \mathcal{L}_{\text {action}}\left(I_{1}, I_{2}\right) \label{combine_loss}
\end{equation}
The $I_{1}$ and $I_{2}$ belong to the domain space $\mathcal{T}_I$, and the parameter $\alpha$ is utilized to manage the impact of the distances between the latent encodings on the structure of the latent space.

	\subsection{Folding action module} \label{fam}
	
	The Folding Action Module (FAM) learns a policy $\pi_{\text{a}}$ to compute the actions $a_i$ between each consecutive pair of fabric states. 
	Each action $a_i$ includes two values: a grasp coordinate $g= [x_g, y_g]$ and a place coordinate $p=[x_p, y_p]$. Given the shortest sequence of intermediate folding steps from the start to the goal state, denoted as $\mathcal{G}$: $\langle I_{\text {0}} =I_{\text {start}},  {I}_{\text {1}}, {I}_{\text {2}}\cdots  {I}_{\text {n-1}}, I_{\text {n}} =I_{\text {goal}} \rangle$, FAM can generate the grasp-and-place coordinates between these states.
	
	For example, after executing the action $a_i$, the fabric can change its state from $I_{\text{i}}$ to $I_{\text{i+1}}$. By computing the optimal grasp-and-place coordinates for each action in the sequence, FAM enables the robot to execute the folding plan efficiently and accurately.
	
	\begin{equation}
		a_{i}=\pi_\text{a}\left(I_{\mathrm{i}}, I_{\mathrm{i}+1}\right)
	\end{equation}
	
\begin{figure}[htbp]
		\centerline
		{
			\includegraphics[
			width=0.6\textwidth]{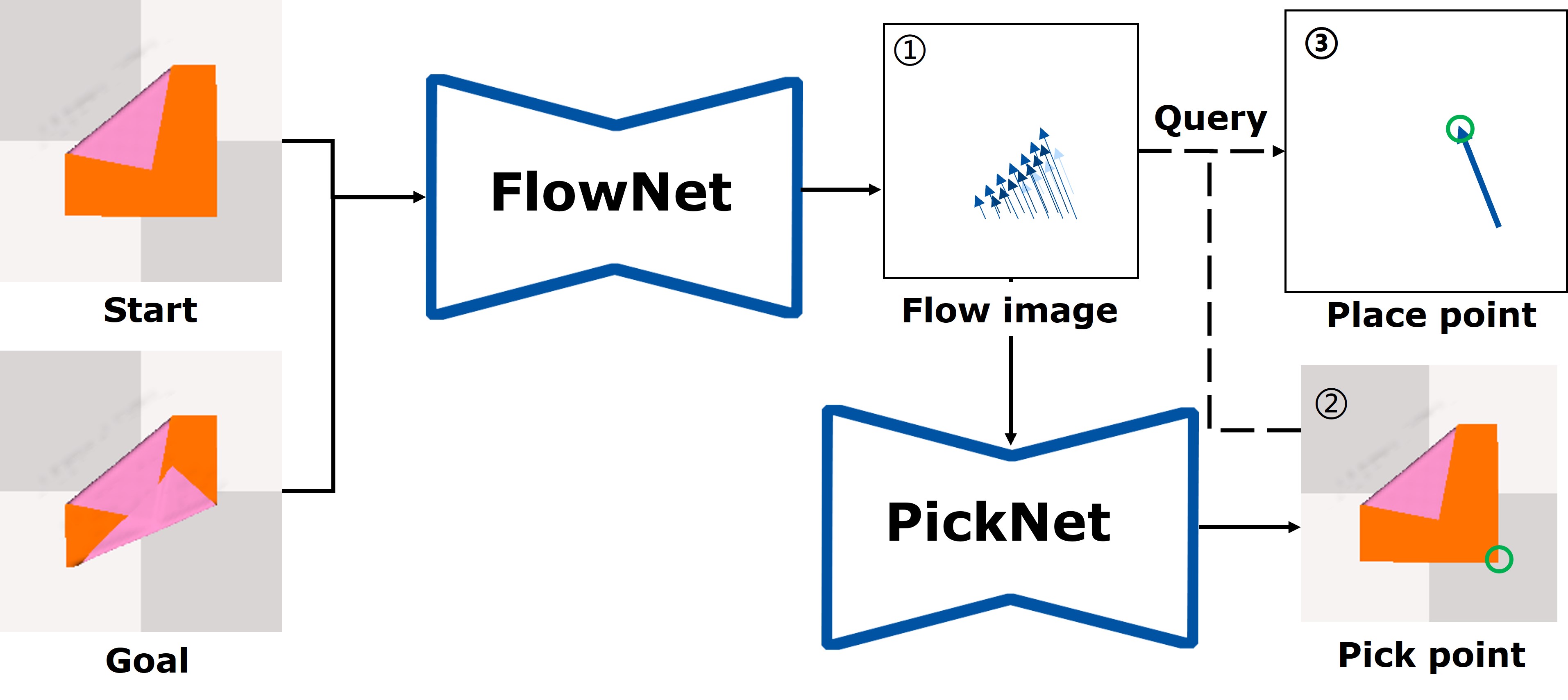}
		}
		\caption{
			The Folding Action Module (FAM) utilizes a flow policy to compute the grasp-and-place points and consists of a FlowNet and a PickNet. The FlowNet is responsible for computing the optical flow of each particle of the fabric, which provides information on how the fabric moves and changes shape during the folding process. The PickNet then reasons about the optimal pick point based on the computed optical flow. Finally, the place point is determined by querying the flow arrow of the pick point, which provides information on the direction and distance of the flow. 

		}
		\label{fig-flownet}
	\end{figure}

	Weng et al.  \cite{weng2022fabricflownet} have demonstrated that computing the grasp-and-place points based on a flow policy can significantly improve the performance of the fabric folding process. To this end, the FAM comprises two main components: a FlowNet and a PickNet, as shown in Figure~\ref{fig-flownet}.
	
	
	In FlowNet, the network's training loss is the endpoint error (EPE), which is the average Euclidean distance between the true flow $f$ and the predicted flow vectors $\hat{f}$ across all pixels.
	Then the loss function of FlowNet is:
	\begin{equation}
		\mathcal{L}_{\mathrm{Flow}}=\frac{1}{N} \sum_{i=1}^{N}\left\| \hat{f}-f\right\|_{2}
	\end{equation}
	
	PickNet is a fully convolutional network that takes the flow image $f$ as input and produces a single heatmap $\hat{H}$ to predict the optimal pick points for the robot arm. The pick point is calculated using $g=arg \, max_g \hat{H}(g)$.
	PickNet is trained using binary cross-entropy (BCE) loss that measures the difference between the predicted heatmaps $\hat{H}$ and the ground truth heatmaps $H$.
	The loss function is:
	\begin{equation}
		\mathcal{L}_{\text {Pick }}=\operatorname{BCE}\left(H, \hat{H}\right) 
	\end{equation}
	
	\subsection{Iterative interactive module}
	
\begin{figure}[htbp]
		\centerline
		{
			\includegraphics[
			width=0.38\textwidth]{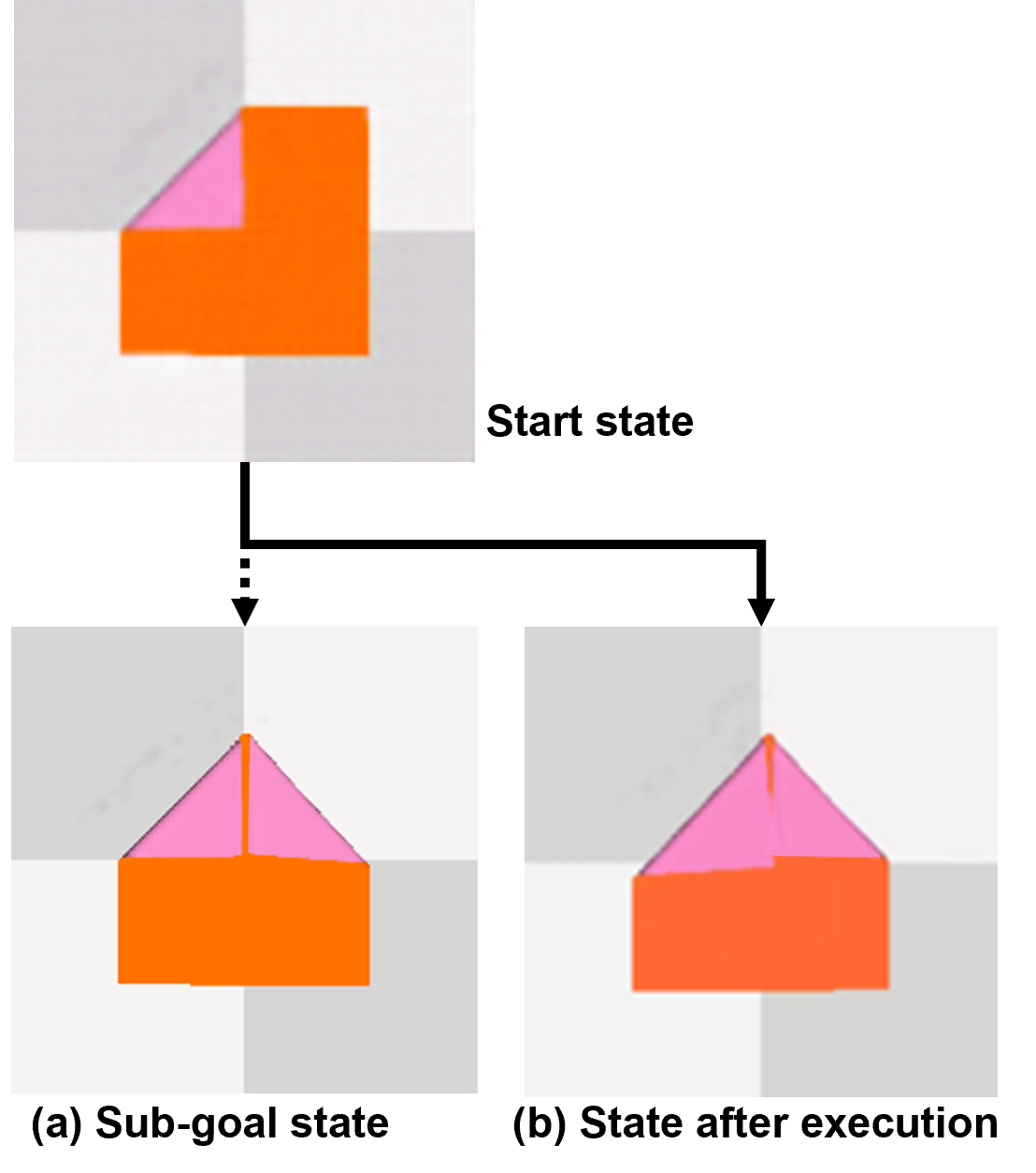}
		}
		\caption{
			Picture (a) shows an inferred intermediate folding sub-goal generated by the FPM, while picture (b) shows the actual state achieved after a grasp-and-place action. As shown, there is a slight visible difference between the two, indicating that there may be deviations between the inferred sub-goals and the actual achieved states.
		}
		\label{fig-difference}
	\end{figure}

	The Iterative Interactive Module (IIM) is a module designed for performing iterative execution of the Folding Planning Module (FPM) and Folding Action Module (FAM) after each grasp and place action until the goal state is achieved in multi-step folding tasks. 
	Due to the deformability of fabric and potential execution uncertainties, the actual execution results may deviate from the goal state. 
	As shown in Figure \ref{fig-difference}, even after executing a single action, there may be a slight difference between the goal state and the resulting state of the fabric. 
	In multi-step folding tasks, these deviations can accumulate and amplify, leading to significant errors.
	
	To address this issue, the IIM is introduced into long-horizon folding tasks. 
	The IIM can decrease the possible uncertainties after each action execution and facilitate more reliable perception and planning, ultimately improving the performance of multi-step tasks. 
	As shown in Figure \ref{fig-framework}, after each action, the IIM considers the new observation of the fabric as a new start state, which is then input into the FPM to generate a new folding plan and infer a new sequence of intermediate states. These intermediate states are then input into the FAM, which calculates and executes a new action, generating a new fabric state. This iterative process continues until the final goal state is reached. 
	For example, in the illustration shown, four iterative steps were performed to reach the goal state.
	
	Through this iterative process, the IIM can decrease execution deviations and improve the efficiency and accuracy of multi-step folding tasks. The IIM is a crucial component in addressing the potential deviations from the goal state and enhancing the overall performance of long-horizon folding tasks.

	\section{Experiment}
	\subsection{Training dataset}

	The training dataset $\mathcal{T}_{I}$ is comprised of tuples in the form $\langle I_{\text {0}}, I_{\text {1}}, \rho=(a,u) \rangle$. 
	In each tuple, $I_{\text {0}} \subset \mathcal{I}$ represents an image corresponding to the initial state, while $I_{\text {1}} \subset \mathcal{I}$ is an image corresponding to the subsequent state. 
	The $\rho$ represents the action that takes place between the two images, where $a\in\left\{0,1\right\}$ indicates whether or not a movement of fabric occurred. 
	The $u$ contains the grasp and place coordinate points of the gripper to effect the change from the state represented by $I_{\text {0}}$ to the state represented by $I_{\text {1}}$.
	Therefore, the tuple $\langle I_{\text {0}}, I_{\text {1}}, (1,u) \rangle$ is an action pair, while the tuple $\langle I_{\text {0}}, I_{\text {1}}, (0,u) \rangle$ is a no-action pair.
	In the no-action pair, the two images are different observations of the same underlying state, where only slight perturbations exist between the two no-action images. 
	
\begin{figure}[htbp]
		\centerline
		{
			\includegraphics[
			width=0.68\textwidth]{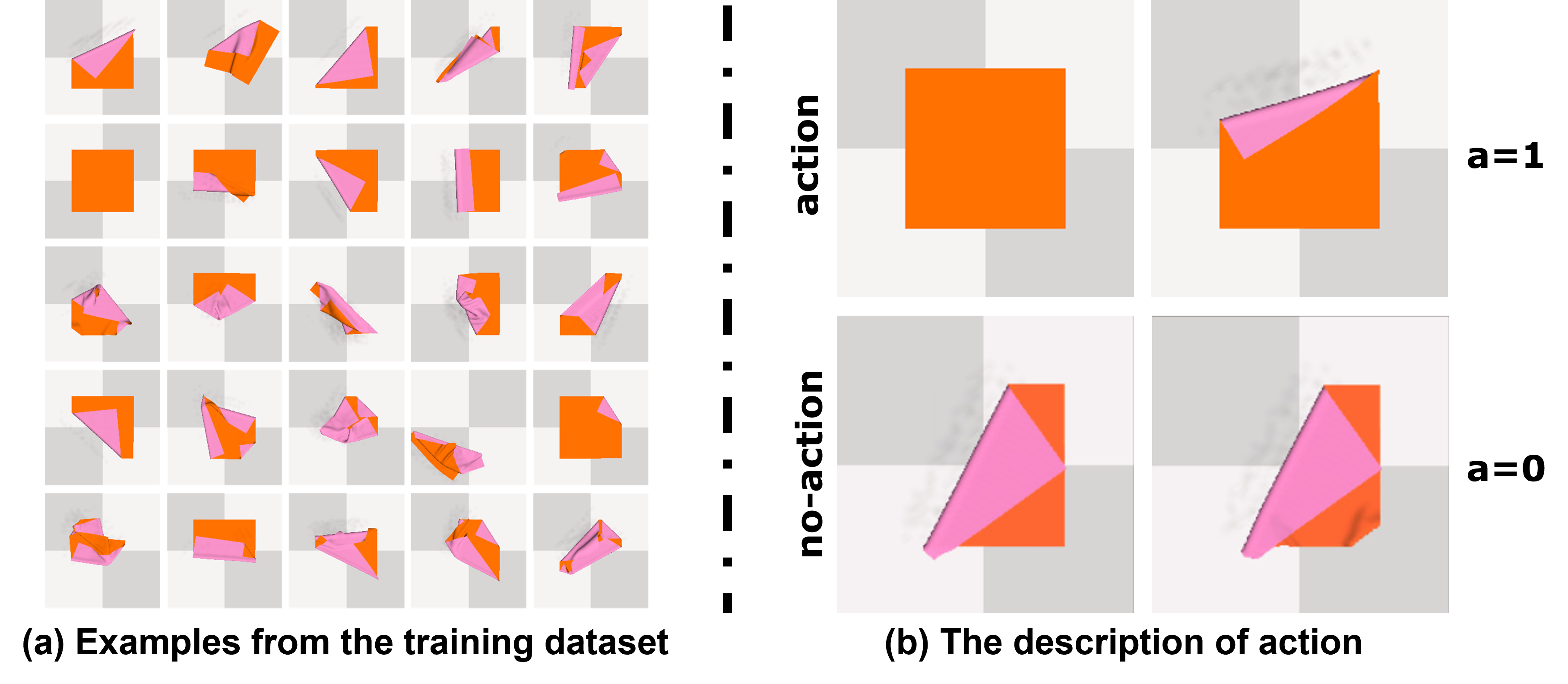}
		}
		
		\caption{
			The (a) are examples from our training dataset. For each sample, we save the initial image, the pick-and-place pixel locations, the next image after action, and the fabric particle positions of both observations. The (b) describes the example of action and no-action pairs in the folding task.
		}
		
		\label{fig-dataset}
	\end{figure}

	To differentiate between no-action and action pairs, we define the action as 0 or 1 based on the particles of fabric, where a=1 indicates that the maximum movement of the fabric particles is beyond the defined movement range. 
	As shown in the right side of Figure \ref{fig-dataset}, in the action pair, the two images exhibit a visible change, and the low-dimensional representations in the latent space are also different. 
	In contrast, in the no-action pair images, the two images exhibit only slight visible changes, and the maximum movement of particles is less than the defined threshold. 
	As a result, their low-dimensional representations in the latent space are the same. 
	
	
	We collect the data for our study using SoftGym, an environment for fabric manipulation built on the particle-based simulator Nvidia Flex. We generate the dataset by taking grasp-and-place actions on the fabric  \cite{weng2022fabricflownet,lippi2022enabling}. Figure \ref{fig-dataset} (left) provides a glimpse of the dataset, which records the fabric's information before and after each action. 
	
	We define the encoded training dataset $\langle z_{\text {0}}, z_{\text {1}}, \rho \rangle \in\mathcal{T}_{z}$ as a set of latent tuples obtained by encoding the input tuples $\langle I_{\text {0}}, I_{\text {1}}, \rho \rangle \in\mathcal{T}_{I}$ into the latent space $\mathcal{Z}_{sys}$. Here, $I_{\text{0}}$ and $I_{\text{1}}$ are the input images,  $z_{\text{0}}$ and $z_{\text{1}}$ are the corresponding encoded latent representations of the images in the latent space $\mathcal{Z}_{sys}$.
	The obtained states $\mathcal{T}_z = \{ z_0, z_1, \cdots , z_m \} \subset  \mathcal{Z}_{sys}$ are called covered states. 
	The approach is data efficient since we utilize a deep learning method that only requires the dataset $\mathcal{T}_{I}$ to contain a subset of possible action pairs of the system that sufficiently cover the dynamics rather than all of them.
	
	\subsection{Implementation details}
	\subsubsection{Network implementation details}
	
	The training datasets, collected by SoftGym \cite{lin2021softgym}, include RGBD images, particle information of the fabric, coordinates of the grasp-and-place action, and masks of the fabric in images.
	In the FPM module, the central network structure of the Variational Autoencoder (VAE) is ResNet. 
	The training of the FPM is carried out for 800 epochs. 
	During training, a batch\_sizes of 32 is used, and the workload is distributed among 8 workers.

	In the FAM module, the FlowNet is similar to the one in  \cite{dosovitskiy2015flownet}, with some changes in terms of the network channels. 
	Additionally, the PickNet, a deep neural network with a depth of six, is employed in this module. 
	The optimization of both modules is performed using the Adam optimizer.
	We set the batch\_sizes of the FAM module as 32 with 6 workers, training epoch is 300.

	The training process is conducted on a machine that runs Ubuntu 18.04. The machine is equipped with a 3.50 GHz Intel i9-9900X CPU, four NVIDIA GTX 2080 Ti GPUs, and 64 GB of RAM.
	In total, the training process takes approximately 57 hours to complete.
	
	\subsubsection{Physical implementation details}
\begin{figure}[htbp]
		\centerline
		{
			\includegraphics[
			width=0.48\textwidth]{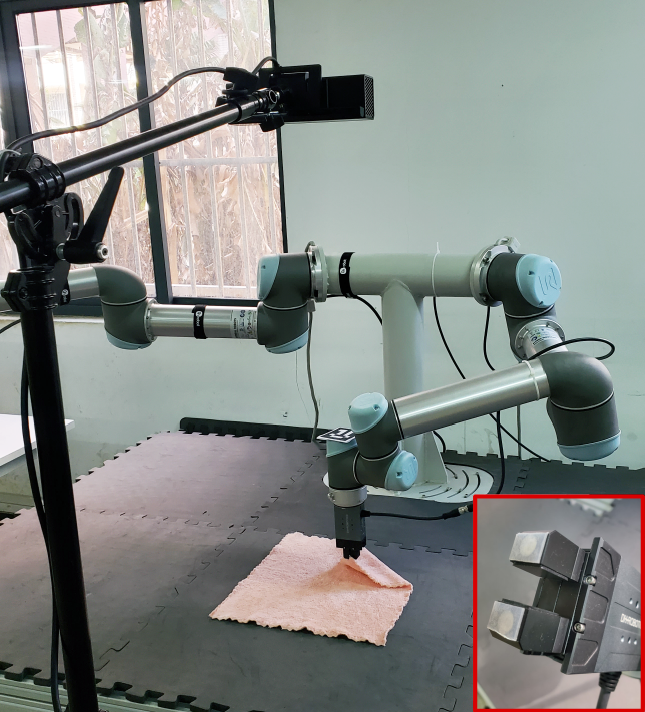}
		}
		\caption{
			In the scene of our physical experiments, we coverd the workspace with several cushions to protect the gripper. To increase the friction between the gripper and the fabric, we paste a non-slip silicone pad on the end of the gripper, as shown in the red box.
		}
		\label{workspace}
	\end{figure}

	We perform the experiments on a 6DOF UR5 robot with DH-Robotics Adaptive Gripper PGE-50-26. The experimental workspace is a 1.7*1.2 meter area. To protect the gripper, we cover the workspace with several cushions. To obtain RGB-D images, we mount a Microsoft Kinect V2.0 sensor 0.9 meters above the workspace. The experimental scene is built into the Robot Operating System (ROS) system  \cite{mittler2017ros} to facilitate the work. Figure \ref{workspace} shows the scene of our physical experiments.
	
	\subsection{Simulation experiments} 
	\subsubsection{Experiment setup}

	We evaluate our DeFNet and compare it to three baselines in the SoftGym environment. We assume that the initial fabric is unfolded and ready to be folded into the goal state. 
	The evaluation metric used is the Mean Particle Distance Error, computed for each fabric particle position coordinates between the achieved and goal fabric configuration.
	We compare our method to the work of Lee et al.  \cite{lee2021learning}, which is a model-free approach. We also compare to Fabric-VSF  \cite{hoque2020visuospatial}, which learns a visual dynamics model with Cross-Entropy Method (CEM). In this experiment, we use the RGB-D dataset as the input to Fabric-VSF. Another baseline approach is LSR-V2  \cite{lippi2022enabling}. A difference between LSR-V2 and our method is that LSR-V2 uses the Action Proposal Module (APM) to predict grasp-and-place points, while we use the Flow-Based Policy.
	For these experiments, we only input the initial and goal states of the fabric to compare the ultimate folded result of all algorithms.
	
	\begin{figure}[htbp]
		\centerline
		{
			\includegraphics[
			width=0.48\textwidth]{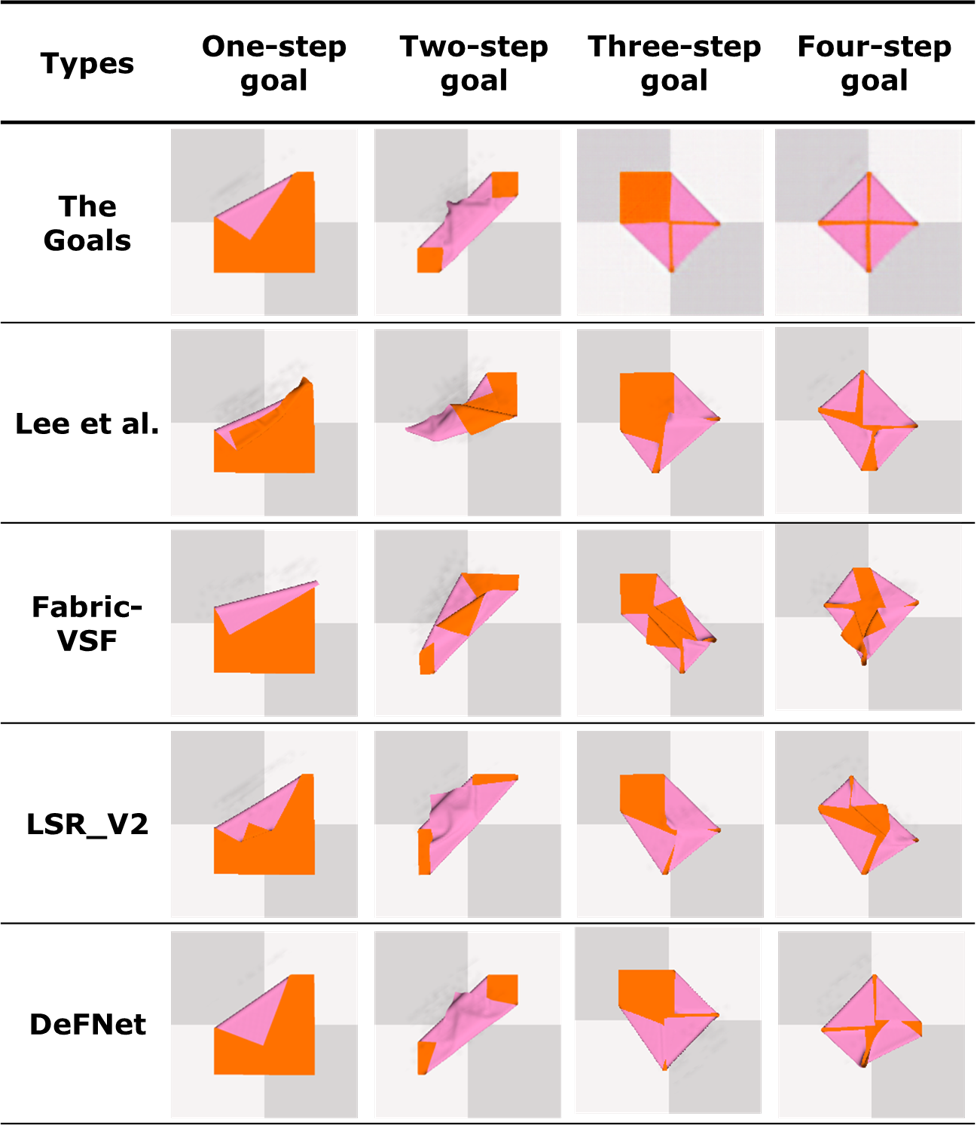}
		}
		\caption{A comparison of manipulation in the simulator. Our approach achieves the best performance in the folding task.}
		\label{fig-result}
	\end{figure}

	\begin{table}[htbp] 
		\centering
		\renewcommand\arraystretch{1.5}
		\tbl{The comparison with baselines. Mean Particle Distance Error (mm) on Fabric Folding Goals. }
		{\begin{tabular}{c c c c c c }
				\hline
				Methods & One-Step & Two-Step & Three-Step & Four-Step  & All \\
				\hline	
				\makecell[c]{Lee et al. \\ 1-Arm \cite{lee2021learning}} & 4.92$\pm$0.51 & 13.83$\pm$1.51& 11.32$\pm$1.32 & 14.27$\pm$1.86 &11.09$\pm$1.42\\
				Fabric-VSF \cite{hoque2020visuospatial} & 3.93$\pm$0.72 & 11.44$\pm$1.16 & 15.71$\pm$1.93 & 19.97$\pm$2.20 &12.76$\pm$1.66\\
				LSR-V2 \cite{lippi2022enabling} & 4.27$\pm$0.43 & 7.08$\pm$1.11 & 12.47$\pm$1.70 & 15.90$\pm$1.33 &9.81$\pm$1.23\\
				DeFNet 
				& \pmb{3.71$\pm$0.32} & \pmb{5.45$\pm$0.58} & \pmb{8.05$\pm$0.90}& \pmb{10.13$\pm$1.20} & \pmb{7.12$\pm$0.86}\\
				\hline	
		\end{tabular}}
			\label{tab-compare}
	\end{table}

	\subsubsection{The result of the simulation experiments} \label{trick}
	Figure~\ref{fig-result} showcases the simulation results of all methods, which are applied to plan identical goals from the same initial fabric state. To assess the folding performance of these methods, we categorize the goal configurations into four tiers, each representing a distinct goal shape that necessitates folding the fabric at least 1, 2, 3, or 4 times. 
	We conducted the comparisons ten times and reported the average particle distance error (in mm) of the fabric.

	\textbf{Mean Particle Distance Error} 
	Based on the four lines in Table~\ref{tab-compare}, it is evident that our method utilizing the deconstructed strategy outperforms the baseline approaches in terms of performance. The mean particle distance error increases for all methods as the folding complexity becomes more challenging. However, our DeFNet stands out with a mean error of 7.12 across all goals, showcasing its exceptional performance.
	
	\begin{table}[htbp]
		\centering
		\renewcommand\arraystretch{1.5}
		\tbl{The number of grasp-and-place actions in the four types of tasks.}
		{
			\begin{tabular}{c c c c c c }
				\hline
				Methods & One-Step & Two-Step & Three-Step & Four-Step \\
				\hline
				Lee et al. 1-Arm \cite{lee2021learning} & 1.0 & 2.0& 3.0 & 4.0 \\
				
				Fabric-VSF \cite{hoque2020visuospatial} & 2.0 & 5.0& 7.0 & 9.0\\
				
				LSR-V2 \cite{lippi2022enabling} & 1.0 & 2.0& 3.0 & 4.0\\
				
				DeFNet & 1.0 & 2.0& 3.0 & 4.0\\
				\hline
			\end{tabular}
		}\label{action-number}
	\end{table}

	\textbf{The Number of Grasp-and-place Actions}	
	We conduct an analysis of the number of grasp-and-place actions performed by each method during the folding tasks. The results, presented in Table~\ref{action-number}, indicate that our method, along with the Lee and LSR-V2 baselines, requires a minimal number of folding actions in each task, with each action corresponding to a single folding step. In contrast, the Fabric-VSF baseline, as reported by Hoque et al.  \cite{hoque2020visuospatial}, requires a larger number of actions. This is likely due to the need for multiple actions when folding a corner, as claimed in their study.

	\subsection{Real world experiments}	
	\begin{figure}[!htbp]
		\centerline
		{	\includegraphics[
			width=0.48\textwidth]{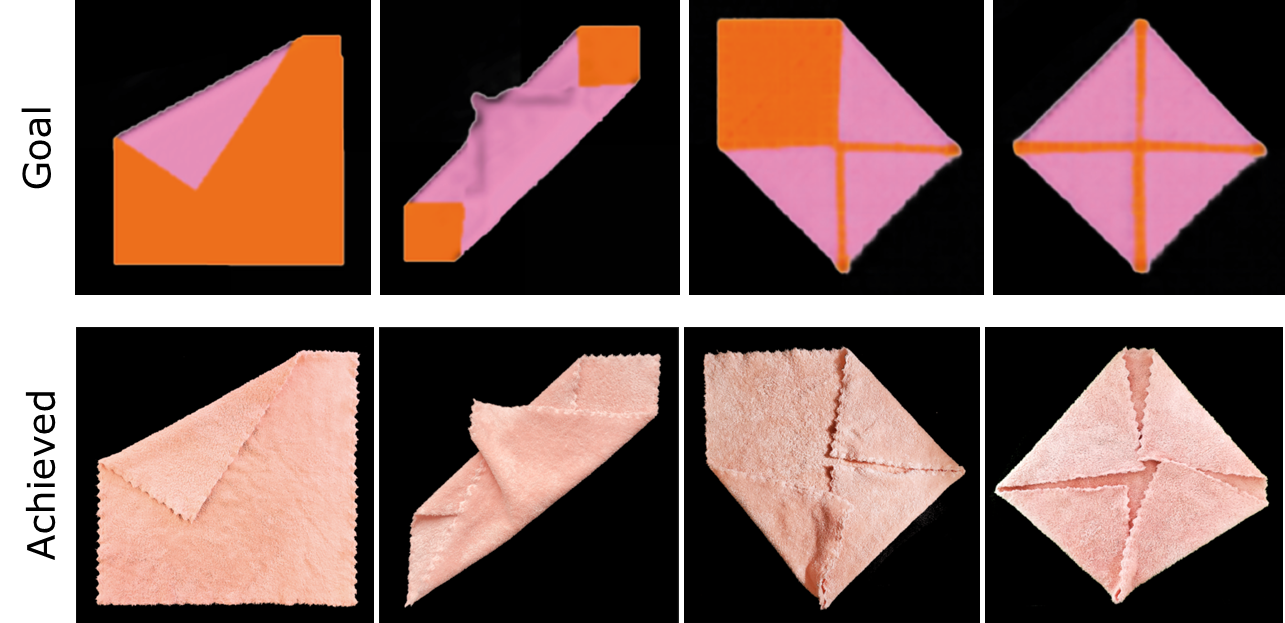}
		}
		\caption{Results for DeFNet on real-world experiments.}
		\label{fig-realresult}
	\end{figure}
	
	\begin{table}[!htbp]
		\centering
		\renewcommand\arraystretch{1.5}
		\tbl{Mean Intersection over Union (MIoU) Over All Fabric Folding Goals in Real World.}
		{
			\begin{tabular}[!htbp]{c c c c c c}
				\hline
				Methods & 	One-Step	& Two-Step 	& Three-Step & Four-Step  \\
				\hline
				DeFNet & 	0.952		& 	0.887 	& 0.921	& 0.826 	\\
				\hline
			\end{tabular}
		}\label{miou}
	\end{table}
	
	Figure~\ref{fig-realresult} illustrates the successful application of our system to a real robot in four folding tasks. While there is a slight disparity between the simulator and the real-world results due to variations in the precision of the mechanism and fabric configurations, the achieved towel closely resembles the goal with minimal visible differences, such as wrinkles and slight biases in details.
	
	Table~\ref{miou} shows the quantitative results. In the real world, obtaining the precise position error of each fabric particle is challenging. Instead, we rely on the mean Intersection-over-Union (MIoU) calculated from the generated fabric masks as a reasonable metric. In the case of One-Step folding tasks, DeFNet achieves a mean IoU of over 0.952, while the Four-Step folding tasks achieve a mean IoU of 0.826.
		
	\section{Ablations}
	
	We perform ablations on our method to evaluate the importance of the three modules of our approach. The ablations are designed to answer the following questions:
	\begin{itemize}
		\item What is the benefit of using a flow policy to determine the folding action in fabric tasks? In this ablation, we entirely apply the method of Lippi \cite{lippi2022enabling} to folding tasks. 
		Lippi utilizes the Action Proposal Module (APM) to determine grasp-and-place points, which is similar to the traditional approach of simultaneously reasoning about both grasp and place points. In contrast, our approach leverages Flownet to compare each pair of the current observation and sub-goal state. The PickNet then utilizes the flow information to reason about the pick point and query the corresponding place point. 
		\item What is the benefit of finding folding paths? In this ablation experiment, we directly input the initial and the final goal states into the flow-based network, FabricFlowNet (FFNet) \cite{weng2022fabricflownet}, without path planning to evaluate the benefit of folding planning in latent space.
		\item What is the benefit of the Iterative Interactive Module (IIM)? We apply the DeFNet without the iterative module, which means the folding actions are all calculated and executed entirely based on the first inferred intermediate folding state.
	\end{itemize}

	\begin{table}[htbp]
		\centering
		\renewcommand\arraystretch{1.5}
		\tbl{Mean Particle Distance Error (mm) Over All Fabric Folding Goals for Ablations.}
		{
			\begin{tabular}[htbp]{c c c c c c}
				\hline
				Methods & One-Step & Two-Step & Three-Step & Four-Step  & All\\
				\hline
				LSR-V2 \cite{lippi2022enabling} & 4.27$\pm$0.43 & 7.08$\pm$1.11 & 12.47$\pm$1.70 & 15.90$\pm$1.33 &9.81$\pm$1.23\\
				
				FFNet without sub-goals  \cite{weng2022fabricflownet}  & 3.69$\pm$0.23 & 24.11$\pm$2.44 & 27.37$\pm$3.84 & 35.54$\pm$4.12& 22.74$\pm$3.31\\
				
				DeFNet without IIM& 3.71$\pm$0.32 & 6.95$\pm$0.64 & 9.33$\pm$0.98& 11.98$\pm$1.23 & 7.98$\pm$0.92\\
				
				DeFNet & 3.71$\pm$0.32 & 5.45$\pm$0.58 & 8.05$\pm$0.90& 10.13$\pm$1.20 & 7.12$\pm$0.86\\
				\hline
			\end{tabular}
		}	\label{ablation}
	\end{table}

	To quantify the impact of the ablations, we conducted additional experiments in the simulator. The results of the first ablation experiments are presented in Section~\ref{trick} and directly utilized in our analysis. The final results of the ablations are summarized in Table~\ref{ablation}.
	
	Comparing the performance of LSR-V2 to the DeFNet without IIM, we observe an 18.7$\%$ improvement when utilizing a flow policy to determine the folding actions. This demonstrates the benefits of employing a flow policy as a representation for actions in fabric tasks.
	
	Furthermore, incorporating latent space to map folding paths led to a significant performance improvement, with the distance decreasing from 22.74 (FFNet without sub-goals) to 7.98 (DeFNet without IIM). The improvement is attributed to the ability of latent space to capture intermediate folding steps. In contrast, the FFNet without sub-goals degenerates into a single-step folding algorithm due to the absence of predicted intermediate folding states.
	
	Lastly, comparing the distance of 7.12 (DeFNet) to 7.98 (DeFNet without IIM), the application of the IIM to DeFNet resulted in a 10.8$\%$ performance improvement. The iterative refinement provided by the IIM reduces uncertainty in predicting the folding process, leading to more accurate results.
	
	
	\section{Conclusion and future work}
	In this work, we introduce DeFNet, a novel approach for fabric folding that comprises three key modules: a Folding Planning Module, a Folding Action Module, and an Iterative Interactive Module. 
	In order to assess the effectiveness of our approach, we carry out experiments on multi-step fabric folding tasks in simulation and compare its performance against three baseline methods.
	We also perform ablation studies to assess the relative contributions of the three modules. 
	Additionally, we apply our approach to an existing robotic system and demonstrate its performance. In future work, we plan to explore the applicability of our method to dual-arm folding tasks.

\bibliographystyle{tfnlm}
\bibliography{reference}
\end{document}